\documentclass[sigconf, manuscript]{acmart}

\AtBeginDocument{%
  \providecommand\BibTeX{{%
    \normalfont B\kern-0.5em{\scshape i\kern-0.25em b}\kern-0.8em\TeX}}}
\setcopyright{acmcopyright}
\copyrightyear{2018}
\acmYear{2018}
\acmDOI{XXXXXXX.XXXXXXX}

\acmConference[Conference acronym 'XX]{Make sure to enter the correct
  conference title from your rights confirmation emai}{June 03--05,
  2018}{Woodstock, NY}
\acmPrice{15.00}
\acmISBN{978-1-4503-XXXX-X/18/06}

\begin{document}

\title{Modelling prospective memory and resilient situated communications via Wizard of Oz}

\author{Yanzhe Li}
\email{Y.li-42@tudelft.nl}
\affiliation{%
  \institution{Delft University of Technology}
  \country{Nederland}
}

\author{Frank Broz}
\affiliation{%
  \institution{Delft University of Technology}
  \country{Nederland}}
\email{F.Broz@tudelft.nl}

\author{Mark Neerincx}
\affiliation{%
  \institution{Delft University of Technology}
  \country{Nederland}
  \email{M.A.Neerincx@tudelft.nl}
}

\renewcommand{\shortauthors}{Trovato and Tobin, et al.}




\maketitle
\pagestyle{plain}

\section{Introduction}
In recent years, there has been an increasing interest in socially assistive robots (SAR), which are intended to become a part of our everyday lives, providing services such as elder care, education, and healthcare. Many of these services necessitate that the robots can communicate and interact with users. However, to interact with robots fluently can be a challenge, for reasons such as an inappropriate mental model of the robot \cite{schramm_warning_2020}. For example, based on the wide range of types of sensors of the robot (e.g., camera, radar, Infra-red detector, microphone, speaker, etc.), users may assume they have multimodal communication capabilities and even expect that the robot can remember them and recall details of their previous interactions \cite{rickert_integrating_2007}. Furthermore, speech may have a variety of accents, dialects, grammatical faults, and disfluencies, making interaction more difficult. This is one of the reasons why much research on human-robot interaction (HRI) has focused on short-term interactions and relied on Wizard of Oz or “constrained rule-based” methods to get around these issues \cite{glas_personal_2017, lee_personalization_2012}. However, communication failures may also happen in these kinds of sets up, especially with elderly participants. 

This abstract presents a scenario for human-robot action in a home setting involving an older adult and a robot. The scenario is designed to explore the envisioned modelling of memory for communication with a SAR. The scenario will enable the gathering of data on failures of speech technology and human-robot communication involving shared memory that may occur during daily activities such as a music-listening activity. The study will use a Wizard of Oz design to collect multimodal communication data about conversation and its context in order to design a model of prospective memory for HRI.

\section{Background}
Prospective memory (PM) is the ability to remember to carry out an intended action at some point in the future \cite{ramos_prospective_2020}. It is an essential cognitive function that allows individuals to plan and execute tasks that are not immediately relevant but will be important in the future. 

Prospective memory is particularly important for older adults because it can help them maintain their independence and quality of life \cite{woods_does_2015}. Prospective memory failures are a common type of memory complaint for older adults \cite{ramos_designing_2016} which can have an impact on their reputation and self-esteem \cite{cohen_prospective_2017} because someone who consistently remembers things is seen as reliable and well-organized, whereas someone who forgets things occasionally is seen as unreliable and disorganized \cite{cohen_prospective_2017}. Thus some individuals rely on external memory aids such as notebooks, diaries and calendars to supplement their memory capacity \cite{mcgee-lennon_user-centred_2011}. The use of digital technology such as smartphones in memory aids designed for individuals is also on the rise \cite{scullin_using_2022}. However, this kind of new technology can be challenging for older adults. An important issue with memory aids is that they can only assist users with prospective memory tasks in a conformist manner, which means that unnecessary reminders may occur due to the lack of situated interaction. To address this issue, we propose the use of a social robot to provide support through a shared prospective memory with the user.

\section{Experimental scenario}

The scenario describes repeatable daily activities where an elderly person is accompanied by a social robot during leisure time. The social robot has a shared memory with the user which can assist the user in daily life.
 
In the proposed scenario, the elderly user and the social robot are engaged in a shared activity of listening to music. These types of leisure activities are likely to be interrupted by other activities of daily living. People would like to be able to switch seamlessly from one activity to the other, but problems with prospective memory make this difficult. The elderly user may, for example,  wish to make tea for himself while listening with the robot. However, due to possible impairments of prospective memory, the elderly user may forget that he had made tea and the kettle will get cold. In this situation, the robot can have shared prospective memory with the user to help accomplish this task. 

With an envisioned shared prospective memory model, the robot should be able to remind the user about the tea when and (only) if a reminder is needed. For example, if the robot observes that an elderly user has gone to the kitchen to take tea or has mentioned in conversation that they are looking forward to their tea in two minutes, then reminders are not needed. This context helps the robot make decisions about reminders. This multimodal, situated approach can be considered an effective way of supporting elderly users in their daily activities.

\section{Discussion}

This scenario highlights how social robots can be used to assist elderly people in their daily lives. By providing personalized assistance with music-listening activities, these robots can help users maintain their independence and improve their quality of life. Additionally, this scenario demonstrates how shared memory can be used to facilitate communication between humans and robots. We are working on modelling prospective memory as a form of memory useful for assisting with daily activities for the elderly. This presents challenges because failures may occur due to speech recognition failures or due to failures in the robot's communication towards the user based on its internal model of the shared experience. Consequently, a method related to memory modelling was proposed to address the aforementioned issues and will be implemented in future work.

In this scenario, we will start with a Wizard-of-Oz set. In this method, a human operator (the Wizard) simulates the behaviour of an automated system by providing responses to user inputs. The Wizard can “overwrite” or “correct” the speech recognition of the older adult's speech to estimate the situated performance of the speech technology and the corresponding chance of failure in the concerned context. Focusing on this scenario lets us collect data about these conversations and their surrounding context for future modelling.

Upon obtaining the data, a shared prospective memory model can be constructed in future work. The social robot can then utilize this memory system to enhance the interaction between itself and a human by sensing the environment and estimating the task state. This enables the robot to evaluate which speech acts are applicable to the current situation and take the initiative in the dialogue in a sensitive way. The robot’s knowledge about the task can also assist it in providing more pertinent and precise responses. In this way, the robot can help improve performance and reduce failures that occur in human-robot interactions.

In conclusion, social robots have great potential for assisting elderly people in their daily lives by providing personalized assistance with prospective memory tasks. Our proposed method related to memory modelling provides a possible solution for addressing the communication failures that may occur between humans and robots. By using the envisioned shared prospective memory models, social robots can enhance their interactions with humans by providing more accurate and relevant responses to assist in tasks of daily living based on shared models. This method has great potential for improving performance and reducing failures that occur in human-robot interactions.


\bibliographystyle{ACM-Reference-Format}
\bibliography{references}

\end{document}